\documentclass{llncs}
\usepackage{llncsdoc}
\usepackage{times}
\usepackage{helvet}
\usepackage{courier}
\usepackage{url}
\usepackage{color,graphicx}
\usepackage[utf8x]{inputenc}
\usepackage{enumerate}

\title{Cognitive Systems and Question Answering}

\author{Ulrich Furbach\inst{1}, Claudia Schon\inst{1} and Frieder Stolzenburg\inst{2}}
\institute{
}
\institute{
	Universit\"at Koblenz-Landau, 
    \email{\{uli,schon\}@uni-koblenz.de}
 \and
	Harz University of Applied Sciences,
   \email{fstolzenburg@hs-harz.de}
}

\begin{document}
\maketitle

\subsection{Autoren} % jeweils max. 4 Zeilen plus Foto

Prof. Dr. Ulrich Furbach ist Professor für Künstliche Intelligenz an der
Universität Koblenz-Landau. Seine Forschungsgebiete umfassen automatisches
Schließen, Wissensrepräsentation, Frage-Antwort-Systeme und
Kognitionsforschung.

\medskip\noindent Dipl.-Inform. Claudia Schon ist wissenschaftliche
Mitarbeiterin in der Arbeitsgruppe K\"unstliche Intelligenz an der Universit\"at
Koblenz-Landau und arbeitet im Forschungsprojekt RatioLog. Ihre
Forschungsinteressen beinhalten K\"unstliche Intelligenz, Kognition und Logik,
wobei ihr Hauptinteresse im Bereich der Beschreibungslogiken liegt.

\medskip\noindent Prof. Dr. Frieder Stolzenburg ist Professor für
Wissensbasierte Systeme an der Hochschule Harz in Wernigerode und leitet das
Labor Mobile Systeme am Fachbereich Automatisierung und Informatik. Seine
Forschungsinteressen umfassen Künstliche Intelligenz, Logik, Kognition sowie
Mobile Robotik.

\subsection{Kontakt}
Universität Koblenz-Landau\\
Universitätsstr. 1\\
56070 Koblenz\\[\medskipamount]
Tel.: +49\,261\,/\,287-2728\\
URL: http://www.uni-koblenz-landau.de/koblenz/fb4/ifi/AGKI

\begin{abstract}This paper briefly characterizes the field of cognitive computing. As an exemplification, the field of natural language question answering is introduced  together with its specific challenges. A possibility to master these challenges is illustrated by a detailed presentation of the LogAnswer system, which is a successful representative of the field of natural language question answering.
\textbf{
%%%
}

\emph{Keywords:} cognitive computing; natural language processing; question answering systems; theorem proving.
\end{abstract}

\noindent Human computer interaction is a discipline with increasing importance.
Many people spend a lot of time with computers playing games, watching movies
but, of course, also solve problems during their professional activities. This
becomes even more important, the more data and information has to be taken into
account. Indeed, this amount is increasing every day. Big data and open data are
keywords that relate to fields of computer science, where exactly these aspects
are tackled.

This paper briefly describes the term \textit{cognitive computing} and demonstrates that natural language question
answering is an example for this new computing paradigm. In the next section, cognitive computing is discussed. After this, a  brief overview on natural language question answering is given. Then the LogAnswer system is described and finally we
conclude with current extensions and future work.

% !TEX root = indman.tex

\section{Cognitive Computing}\label{sec:cognitivecomputing}

IBM is certainly one of the major companies that pushed the development of
modern computers from the very beginning. With respect to the development of
intelligent machines, IBM succeeded  twice to set a milestone: In 1997 the chess
playing computer \textit{Deep Blue} managed to beat the world-champion Garry
Kasparov. There was a discussion after this match whether the IBM team was
cheating during the tournament. Kasparov demanded a rematch, which was refuted
and, even more,  Deep Blue was dismantled. In 2011 the IBM computer system
\textit{Watson} beat two former  winners in the quiz-show Jeopardy. In
Jeopardy, the players have to understand natural language questions from various
domains and give quick answers. This kind of question answering and reasoning is
called deep question answering. The Watson system used many different sources
of knowledge. Being not connected to the internet, Watson had access to
databases, dictionaries, encyclopedias, formal ontologies but also literary
works and newspaper articles.

Very different to Deep Blue, after this effective public event, the Watson
system was  developed further and also tailored to various application domains
\cite{Watson2013}. It is now applied in eHealth, cancer research, finance and
the list is steadily increasing. There is even a version of Watson which is
acting as chef, creating really extraordinary dishes, e.g. a Vietnamese Apple
Kebab \cite{chef2014}. The keyword which turns the Jeopardy winning system into
the basis of a business plan is \textit{cognitive computing system}. Such a
system is designed to learn and to interact with people in a way that the result
could not be achieved either by humans or machine on their own. Of course,
mastering \textit{Big Data} also plays an important role --  IBM's marketing
slogan is "Artificial Intelligence meets Business Intelligence". Such a
cognitive computing system has the following properties:

%\paragraph{Properties of a cognitive computing system}
\begin{enumerate}[(a)]
\item Multiple knowledge formats have to be processed:  Formal knowledge, like ontologies but also a broad variety of natural language sources, like textbooks, encyclopedias, newspapers and literary works.\label{a}
\item The different formats of knowledge also entail the necessity to work with different reasoning mechanisms, including information retrieval, automated deduction in formal logic and probabilistic reasoning. \label{b}
\item The different parts and modules have to interact and cooperate very closely.\label{c}
\item The entire processing is time critical, because of the interaction with humans.\label{d}
\item The system must be aware of its own state and accuracy in order to rank its outcome.\label{e}
\end{enumerate}

Natural language question answering is obviously one example of cognitive
computing as depicted above. There are one or several  huge text corpora together
with other background knowledge, which can be given in various formats. The user
interaction is rather simple: The user asks a natural language question and the
system answers in natural language. In the following natural language question answering is briefly introduced.
%In the following section we focus on such systems an in particular to our own
%projects \textit{LogAnswer} and \textit{RatioLog}.

\section{Natural Language Question-Answering}
\label{sec:nlp_qa}

% !TEX root = indman.tex

%Beginning in the 1950s, natural language processing focused on the task of
%machine translation. Since this task turns out to be very difficult, later
%systems tried to work in restricted worlds, e.g. so-called blocks worlds in the
%SHRDLU project \cite{Winograd1972}, where a robot plans to build one or more
%vertical stacks of blocks according to natural language instructions. Another
%system was ELIZA, a simulation of a Rogerian psychotherapist \cite{Weizenbaum83}.
%This program simulated human-like interaction by simple pattern-matching
%techniques. Among others, the system turned assertions into questions, e.g.
%responding to \emph{I have problems with my family.} into \emph{Why do you say
%you have problems with your family}.

Up to the 1980s, most systems processing natural language were based on
explicit, hand-coded rules. With these rules, the syntactic structure of
sentences or the dependency of semantic constituents was analyzed. Here,
utterances were understood as to have a syntax characterized by a formal
grammar, in particular, a context-free grammar. Later on, syntactic rules
were enhanced by feature descriptions of syntactic constituents, e.g. in
case grammars or semantic nets \cite{DBLP:series/cogtech/Helbig2006}. These
representations can also be expressed by logical formulas, which lead to
so-called phrase-structure grammars. One of the latest and most prominent
theories in this respect is HPSG -- Head-Driven Phrase Structure Grammar
\cite{PS94}.
%An HPSG grammar includes principles and grammar rules and lexicon entries. These
%logic-based rules helped to identify certain constituents of utterances, e.g.
%its subject or agent, which can be used in natural language systems.

Starting in the 1980s, machine learning (ML) techniques were introduced in
the field of natural language processing (NLP). This changed the field completely.
Until that time, the idea was that in order to process natural language,
the sentential structure of an utterance has to be analyzed and hence
somehow the meaning understood first. But since computers became much
faster, more or less brute-force methods based on statistical analyses and
machine learning were employed. Hidden Markov Models were used to predict
which word or part of speech is used next \cite{DBLP:books/daglib/0001548}.

Nowadays, computers and the World Wide Web provide an ever-growing amount of
digitally stored knowledge, which is accessible to anyone from home, the
workplace or even with mobile devices. While the abundance of
available information offers manifold benefits, it can also make the search for
some particular data quite tedious. The tool of choice is usually a search
engine. However, this is inadequate if the user has a specific question in mind:
Instead of simply entering a question, one has to guess suitable keywords.

The field of question answering (QA) intends to improve this search process. A
QA system communicates with the user in natural language. It accepts
properly formulated questions and returns concise answers. These automatically
generated answers are usually not extracted from the web. Rather, the QA system
operates on an extensive knowledge base which has been derived from textual
sources, employing a natural language interface allowing untrained users an
intuitive interaction with the system.
%Extracted answers are suitable for further automated processing, making QA
%systems suitable for embedding in other knowledge-related applications.

Currently, many
QA systems rely on shallow linguistic methods for answer derivation, however, 
with little attempt to include semantics. This may prevent finding an answer. For
example, a superficial word matching algorithm can fail when the textual sources
use synonyms of the words in the question. Hence a system must model some form
of background knowledge. In summary, cognitive aspects of linguistic
analysis, e.g. semantic nets in a logical representation, should be combined
with machine learning techniques, e.g. when determining the most appropriate
answer candidate -- as done in the LogAnswer system.

\section{The LogAnswer System}
\label{sec:loganswer}
% !TEX root = indman.tex

%\section{The LogAnswer System}

\begin{figure}[top]
\centering
\includegraphics[width=\textwidth]{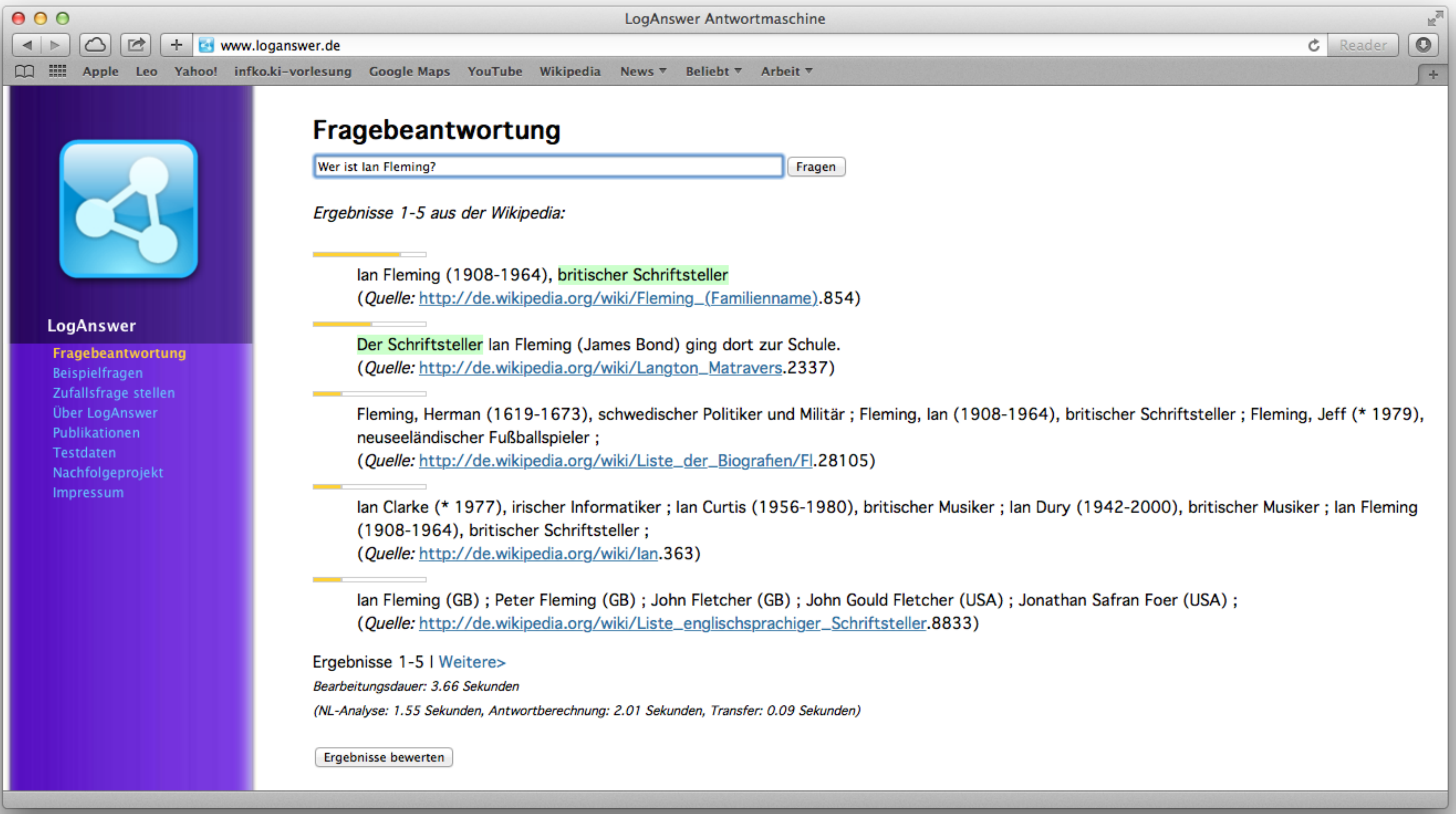}
\caption{Screenshot of the LogAnswer System.}%Hyper
\label{fig:loganswer}
\end{figure}

LogAnswer \cite{Furbach:Gloeckner:Helbig:Pelzer:LogicBasedQA:2009} is an open
domain question answering system. It is accessible by a web interface
(\url{http://www.loganswer.de}) similar to that of a search engine, see Fig.
\ref{fig:loganswer}. The user enters a question into the text box and LogAnswer
presents the three best answers, which are highlighted in the relevant textual
sources to provide a context. 
%The given answers are derived from an extensive knowledge base which was created from a snapshot of the German Wikipedia. 
While many systems for natural language question answering focus on shallow linguistic methods, LogAnswer uses an automated theorem prover (ATP) to compute the replies.

The system was developed in the LogAnswer project which was a cooperation
between the IICS (Intelligent Information and Communication Systems) at the
Fernuniversit\"at in Hagen and the Artificial Intelligence Research Group (AGKI)
at the University Koblenz-Landau. The project was funded by the German Research
Foundation DFG (Deutsche Forschungsgemeinschaft) and aimed at the development of
efficient and robust methods for logic-based question answering. 
%In the LogAnswer prototype \cite{Furbachetal2008} that provides a testbed for
%these approaches, we employ a logical knowledge representation to capture the
%semantics, and answers are extracted using a combination of linguistic
%techniques and automated reasoning. 
The IICS is experienced in computational linguistics and knowledge engineering.
Within the LogAnswer project the IICS handled the natural language aspects and
provided the knowledge base. As an expert in automated theorem proving, the AGKI
was responsible for the deductive aspects of the LogAnswer project.

As indicated in (\ref{c}) in the list of properties of cognitive computing
systems, it is important to take care that the different modules interact and
cooperate closely. When combining NLP and automated reasoning as in the
LogAnswer system, paying attention to the conflicting aims of the two fields is
important. Since NLP methods are often confronted with flawed textual data, they
strive toward robustness and speed. Nevertheless, they lack the ability to
perform complex inferences. In contrast to that, a theorem prover uses a sound
calculus to derive precise complex proofs. However, even minor flaws or
omissions in the data can lead to a failure of the derivation process. 
Furthermore, refutationally complete theorem provers can have problems when
dealing with large amounts of data due to the fact that they can easily get
stuck performing redundant inferences. In the LogAnswer system NLP is used to
filter the input for the theorem prover to a fraction of the knowledge available
to LogAnswer, and the prover is embedded into a relaxation mechanism which can
lessen the proof requirements for imperfect input data
\cite{DBLP:journals/aicom/FurbachGP10}.

As claimed in (\ref{a}) in the list of properties, the LogAnswer system uses
multiple knowledge formats. One part of the knowledge is provided by a snapshot
of the German Wikipedia, which has been translated into a semantic network
representation in the MultiNet (Multilayered Extended Semantic Networks)
formalism \cite{DBLP:series/cogtech/Helbig2006}. To make the semantic networks
accessible to modern theorem provers, LogAnswer is also equipped with a
representation of the MultiNet knowledge base in first-order logic (FOL). See
\cite{DBLP:journals/aicom/FurbachGP10} for details on the translation of the
MultiNet knowledge base into a first-order logic knowledge base. All in all,
29.1 million natural language sentences have been translated. In addition to
that, a background knowledge consisting of 12,000 logical rules and facts is
used. This background knowledge provides general knowledge which is advantageous
for the setting of question answering. Automated reasoning enables the
integration of this background knowledge. 
%The expressivity of MultiNet exceeds that of FOL, and while some aspects of
%MultiNet are lost in the translation, we approximate the expressivity by using
%logic extensions like equality and arithmetic evaluation. See [2] for a detailed
%example of the logical translation and subsequent processing.

\begin{figure}[top]
\centering
\includegraphics[width=\textwidth]{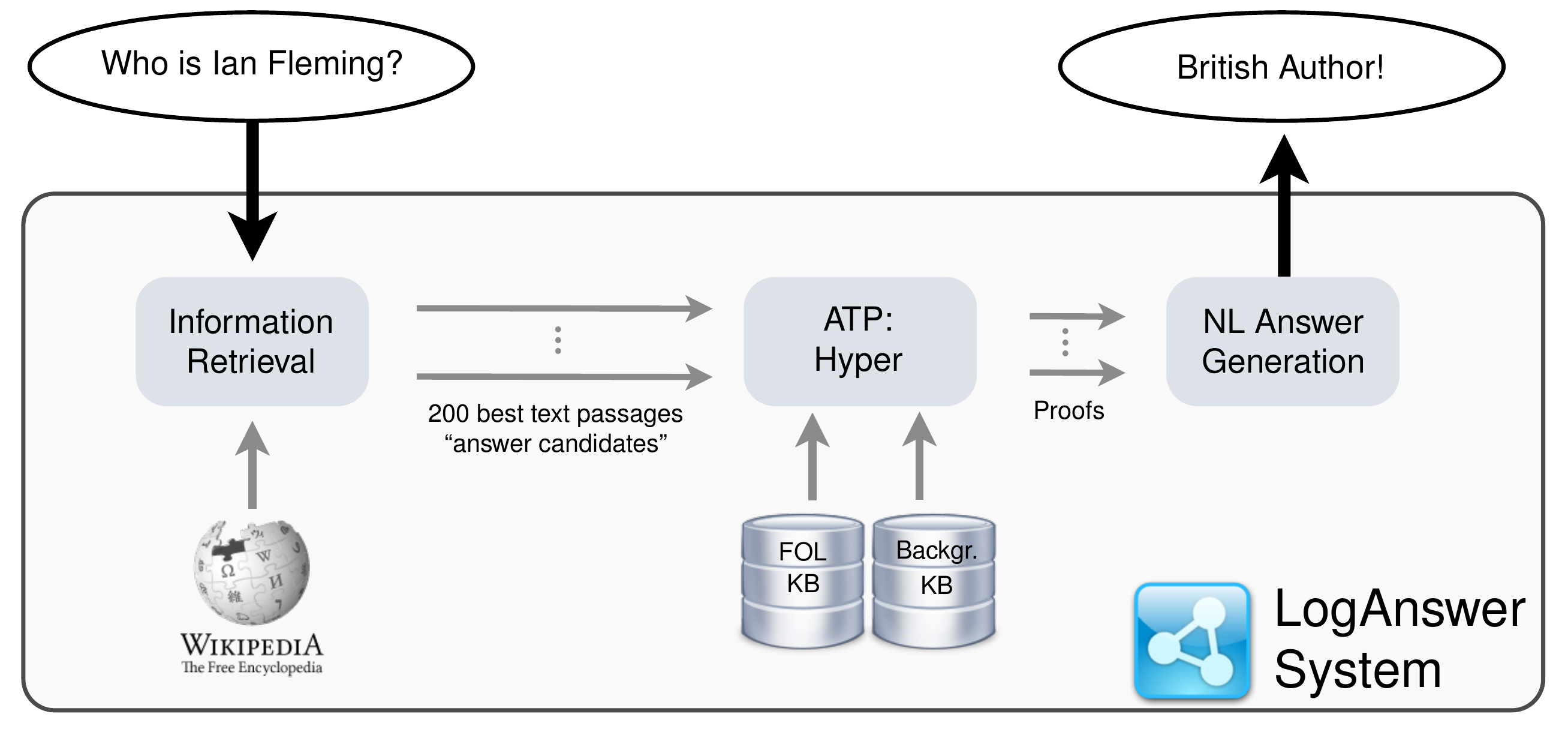}
\caption{Question Processing of the LogAnswer System.}
\label{fig:qa}
\end{figure}

In Figure~\ref{fig:qa} it is depicted how LogAnswer processes a
question. Since it is a web-based question answering system, users expect
the system to respond quickly. This aspect of time criticality
corresponds to (\ref{d}) in the list of properties and is a serious restriction
of the time available for the LogAnswer system to process a question. In such a
restricted time, a question cannot be answered directly using the the whole
knowledge base. Therefore, several different techniques such as natural language
processing, information retrieval, machine learning and automated deduction come
to use. This corresponds to claim (\ref{b}) in the list of properties. After
translating the question into the MultiNet and FOL representation, the
Wikipedia content is matched against the given query using retrieval and shallow
linguistic criteria.  By this, lists of features like the number of matching
lexemes between passages and the question or the occurrences of proper names in
the passage are computed.  Afterwards an ML algorithm ranks
text passages using these features. Then up to 200 text passages are extracted
from the knowledge base according to this ranking. These so-called \emph{answer
candidates} have a high probability to contain the answer and can be computed
rapidly. The computation of feature lists is implemented robustly, which allows
to handle documents containing syntactic errors and thus to extract answers from
text passages which cannot be parsed completely. In the next step the theorem prover Hyper
\cite{WernhardPelzer} is used. The Hyper theorem prover is an implementation of
the hypertableaux calculus \cite{DBLP:conf/jelia/BaumgartnerFN96} extended with
equality. It has been shown to be very suitable for the type of reasoning problems
occurring in the question answering setting, which are characterized by their
large number of irrelevant axioms. 

With the help of Hyper the answer candidates are tested consecutively. For each
of these tests, the logical representation of both the query and an answer
candidate together with the background knowledge are fed into Hyper. A
successful proof provides an answer by giving an instantiation of the variables
of the logical representation of the query.  If no proof
can be found in time, query relaxation techniques come to pass. These techniques
allow certain subgoals of the query to be weakened or dropped in order to enable
the prover to find a proof in short time. Query relaxation increases the
likelihood of finding an answer even if the knowledge at hand is incomplete.
However, the drawback of this technique is that it decreases the probability
that the answer found is relevant to the query. As claimed in (\ref{e}) in the
list of properties, the LogAnswer system is aware of its own accuracy, because
all proofs are ranked by machine learning algorithms. The
three proofs with the highest rank are translated back into natural language
answers and are presented to the user.

\section{Conclusions}
In this paper, the state of the art in cognitive computing systems and in
natural language question answering is discussed. As a prototypical example, the
LogAnswer system is described in detail and its properties are checked against
criteria for cognitive computing systems.

Currently the LogAnswer system is extended in the follow-up project
RatioLog, aiming at the inclusion of rational and human-like reasoning
components. In \cite{kik}  the use of deontic logic for modeling
human reasoning and its automating with the logical machinery from
LogAnswer is demonstrated. Another extension is with respect to defeasible reasoning
\cite{WS14}, which is helpful to determine the best answer from the  possibly contradicting answer candidates.

\bigskip

% Danksagung
\noindent\emph{Dieser Beitrag entstand im Rahmen des Projekts RatioLog --
Rationale Erweiterungen des Logischen Schließens, das von der Deutschen
Forschungsgemeinschaft (DFG) unter den Kennzeichen FU~263/15-1 und STO~421/5-1
gefördert wird.}

\renewcommand{\sc}{}
\bibliographystyle{acm}
\bibliography{wissen}

\end{document}